# Accelerating RNN Transducer Inference via One-Step Constrained Beam Search

Juntae Kim and Yoonhan Lee

*Abstract*—We propose a one-step constrained (OSC) beam search to accelerate recurrent neural network (RNN) transducer (RNN-T) inference. The original RNN-T beam search has a while-loop leading to speed down of the decoding process. The OSC beam search eliminates this while-loop by vectorizing multiple hypotheses. This vectorization is nontrivial as the expansion of the hypotheses within the original RNN-T beam search can be different from each other. However, we found that the hypotheses expanded only once at each decoding step in most cases; thus, we constrained the maximum expansion number to one, thereby allowing vectorization of the hypotheses. For further acceleration, we assign constraints to the prefixes of the hypotheses to prune the redundant search space. In addition, OSC beam search has duplication check among hypotheses during the decoding process as duplication can undesirably shrink the search space. We achieved significant speedup compared with other RNN-T beam search methods with lower phoneme and word error rate.

## I. INTRODUCTION

The success of deep learning technology ushered a new era for automatic speech recognition (ASR) [1–5]. One of recent research streams of ASR is end-to-end (E2E) framework [4–9], which can directly map incoming speech signals into character sequences, thus considerably simplifying complex procedures of traditional ASR systems including acoustics, pronunciations, and language models [1, 2].

The critical evaluation metrics of a practical ASR system are ASR accuracy, streaming capacity, and computationally efficient decoding [9]. Recent E2E ASR methods have already outperformed traditional ASR systems in terms of accuracy using various E2E frameworks based on attention-based encoder-decoder networks (AEDNs) [4, 6] and recurrent neural network transducers (RNN-T) [7–9]. However, some limitations remain concerning decoding efficiency.

Beam search [10] is a search algorithm widely used for the decoding phase, which restricts the search space to reduce computational complexity while finding a sub-optimal path from a given lattice. For each decoding step, hypotheses are expanded only once from their root nodes; these expansions are conducted per hypothesis for a beam search in AEDN-based methods [11], i.e., a loop program for hypothesis traversal is involved in the beam search algorithm. In [11], the authors removed the for-loop by vectorizing the hypotheses, which accelerated the beam search procedure considerably. However, they focused on beam search for AEDN-based methods that cannot be streamed owing to their inherent attention mechanisms that consume the entire utterance.

The RNN-T is a representative streamable E2E framework. Unlike AEDN-based methods, the hypotheses of the beam search for RNN-T can be expanded in an unrestricted manner at each decoding step [7], i.e., the expansion behaviors of the hypotheses are different; thus, vectorizing hypotheses is nontrivial problem. Instead of vectorization, heuristic pruning was proposed to exit the while-loop—within the original RNN-T beam search—earlier to reduce the computational complexity to some degree, while the while-loop still exists [12].

Herein, we propose an accelerated beam search for RNN-T: one-step constrained (OSC) beam search; its main benefits are: (i) Major acceleration achieved by vectorizing hypotheses, i.e., removing the while-loop. This vectorization is possible as we found that the unrestricted expansion can be constrained to expand just once or not at all. (ii) Further improvement in the computational efficiency from pruning the prefixes of the hypotheses. (iii) The word and phoneme error rate (WER, PER) improvement from duplication check as duplication among hypotheses can occur during the RNN-T decoding process, which can degrade WER and PER.

## II. RNN TRANSDUCER

The RNN-T transcribes $\mathbf{x} = \{x_t\}_{t=1}^T$ into $\mathbf{y} = \{y_u\}_{u=1}^U$, where $x_t$ is an acoustic frame, $t$ is the frame index, $y_u$ is an output label and $u$ is the output label index. In general, the length of acoustic frames $T$ is much larger than that of output labels $U$. The key idea of RNN-T to address the length difference between $\mathbf{x}$ and $\mathbf{y}$ is adopting the blank symbol, which allows RNN-T to decide whether to produce the output label during the $T$-step decoding procedure. For instance, one of the decoding paths of RNN-T concerning $\mathbf{y} = [y_1, y_2, y_3]$ can be $\mathbf{y}^* = [y_1, \phi, \phi, \cdots, y_2, \phi, y_3]$, where $\phi$ is the blank symbol, $|\mathbf{y}| = U$ and $|\mathbf{y}^*| = T$. Further details of RNN-T are as follows.

### A. Model Architecture

The framework of the RNN-T comprises the encoder, prediction and joint networks. The encoder network maps the input acoustic frame $x_t$ into the hidden representation $h_t^{enc}$ as

$$h_t^{enc} = f^{enc}(x_t, h_{t-1}^{enc}), \qquad (1)$$

The prediction network estimates the hidden state $h_u^{pre}$ based

TABLE I
THE INVESTIGATION OF EXPANSION AND PREFIX SEARCH

|   | Expansion Search | Prefix Search |
|---|---|---|
| 1 | 97.73 | 84.42 |
| 2 | 2.24 | 13.93 |
| 3 | 0.03 | 1.65 |

In the expansion search, the number of expansions was investigated by comparing $A$ in line 3 and $B$ in line 18. The results were accumulated for all $t$-steps, where $t \in [1, T]$. In the prefix search, for each sequence $\mathbf{y}$ in $A$, length differences between $\mathbf{y}$ and its prefixes $\hat{\mathbf{y}} \in pref(\mathbf{y}) \cap A$ were calculated; the results were accumulated for all the $t$-steps, where $t \in [1, T]$. For better representation, the results for the prefix and expansion search are normalized to the ratio (%). This experiment was conducted on TIMIT dataset [13].

on previous non-blank output label $y_{u-1}$ as,

$$h_u^{pre} = f^{pre}(y_{u-1}, h_{u-1}^{pre}), \quad (2)$$

The joint network combines the hidden representations of the encoder and prediction network as,

$$\begin{aligned} z_{t,u} &= f^{joint}(h_t^{enc}, h_u^{pre}) \\ &= \tanh(W_e h_t^{enc} + W_p h_u^{pre} + b_z) \end{aligned}, \quad (3)$$

where $W$ and $b$ with subscripts are the weight matrix and bias vector, respectively. The posterior probability for each output label $k$ is obtained by applying a softmax layer as follows,

$$h_{t,u} = linear(z_{t,u}) = W_z z_{t,u} + b_s; \quad (4)$$

$$\Pr(k | t, u) = \Pr(k | \mathbf{y}, t) = softmax(h_{k,t,u}), \quad (5)$$

where $\mathbf{y} = [y_1, \cdots, y_u]$.

*B. Beam Search for RNN Transducer*

The beam search—including the prefix (lines 5–7) and expansion searches (lines 8–17)—as described in Algorithm 1 is used for the decoding process of the RNN-T [7]. In this study, $pref(\mathbf{y})$ is the set of prefix sequences of $\mathbf{y}$, $K$ is the set of output label indexes without blanks, and $W$ is the beam width. Further, when updating the posterior probability of the hypothesis, it must be ended with blank probability because in line 11 at each $t$-step, we denote such a posterior probability as *complete* and the other as an *incomplete probability*.

The expansion search in Algorithm 1 is a sequential process that uses the while-loop. This process causes the following problems: (i) As $\mathbf{y}^*$ changes at each iteration of the while-loop—the subject sequence of the expansion is not fixed and depends on the previous iteration—the parallelization of the expansion search is difficult. (ii) As the iterations of the while-loop continue until "$B$ contains less than $W$ elements that are more probable in $A$," the decoding time can be exponentially increased according to $W$.

Further, Algorithm 1 does not check duplication within the set of hypotheses in the beam search procedure, and thus, duplicated hypotheses can be included.

III. ONE-STEP CONSTRAINED BEAM SEARCH FOR RNN-T

Time alignment among the output labels and acoustic frames is almost linear [7]. Thus, an optimal decoding path is highly likely to exist in the region of an output lattice close to its main diagonal. From the intrinsic relationship between output labels and acoustic frames, we assume that there is no need to expand the hypothesis in an unrestricted manner and investigate

---

**Algorithm 1**: Beam Search for RNN-T in [7]

1. **Initialize**: $B = \{\phi\}$; $\Pr(\phi) = 1$
2. **for** $t = 1$ **to** $T$ **do**
3.    $A = B$
4.    $B = \{\}$
5.    **for y in** $A$ **do**
6.       $\Pr(\mathbf{y}) \mathrel{+}= \sum_{\hat{\mathbf{y}} \in pref(\mathbf{y}) \cap A} \Pr(\hat{\mathbf{y}}) \Pr(\mathbf{y} | \hat{\mathbf{y}}, t)$
7.    **end for**
8.    **while** $B$ contains less than $W$ elements more probable than the most probable in $A$ **do**
9.       $\mathbf{y}^* = $ most probable in $A$
10.      Remove $\mathbf{y}^*$ from $A$
11.      $\Pr(\mathbf{y}^*) = \Pr(\mathbf{y}^*) \Pr(\phi | \mathbf{y}, t)$
12.      Add $\mathbf{y}^*$ to $B$
13.      **for** $k \in K$ **do**
14.         $\Pr(\mathbf{y}^* + k) = \Pr(\mathbf{y}^*) \Pr(k | \mathbf{y}^*, t)$
15.         Add $\mathbf{y}^* + k$ to $A$
16.      **end for**
17.    **end while**
18.    Remove all but the $W$ most probable from $B$
19. **end for**
20. **Return:** $\mathbf{y}$ with highest $\log \Pr(\mathbf{y}) / |\mathbf{y}|$ in $B$

---

prefixes that have significant length differences with the corresponding hypothesis at specific decoding steps as shown in Algorithm 1. This assumption is verified by studying the behaviors of expansion and prefix searches in Algorithm 1. Table I summarizes the results, which indicate that not only the number of expansions is 1 in most cases (97.73 %) in the expansion search, but also most length differences between the prefix and its corresponding hypothesis are 1 (84.42 %) in the prefix search at each decoding step.

Thus, based on this study, we assign the following constraints to Algorithm 1: (i) In the expansion search, the possible expansion number can be constrained to 1, referred to as the OSC. Note that OSC for an expansion search facilitates the parallelization of the hypothesis expansion procedure as the OSC leads from irregular expansion behavior to a regular one by expanding once or not at all. (ii) In the prefix search, studying prefixes such that $|\mathbf{y}| - |\hat{\mathbf{y}}| > \alpha$ can be discarded.

Based on these constraints, we proposed an OSC beam search, as described in Algorithm 2. The code consists of constrained prefix search (lines 5–7), expansion search with the vectorized calculation of all probabilities for the possible expansions of the hypotheses (lines 8–10), local pruning (line 11), duplication checks (lines 12), and global pruning (line 13).

In line 1, $\mathbf{0}$ and $\mathbf{0}'$ are zero vectors corresponding to the initial inputs and states, respectively. In line 3, $A_t^l$ is the set of previous hypotheses and $B_t^l$ in line 4 is the set of current hypotheses, updated in line 13. In the constrained prefix search, we simply add a condition to the prefix, $|\mathbf{y}_i| - |\hat{\mathbf{y}}| \leq \alpha$ as in line 6 where $i \in \mathbb{N}$ is the hypothesis index and $\mathbf{y}_i \in A_t^l$.

The expansion search in Algorithm 1 is reformulated by vectorizing the calculation of posterior probabilities with regard to (1)-(5) and by eliminating the while-loop in Algorithm 1 (line 8). The expansion search in Algorithm 2 can be broadly described as calculating the *complete* and *incomplete probabilities* for the hypotheses with no expansions (line 9) and expansions (line 10), respectively. The detailed $t$-step expansion search is described as follows.

To calculate all posterior probabilities for possible expansions of the hypotheses in $A_t^l$, where $\bar{\mathbf{h}}^{pre}$ is first obtained by joining $\mathbf{h}_S^{pre}$ and $\mathbf{h}_V^{pre}$ as

$$\bar{\mathbf{h}}^{pre} = [\mathbf{h}_S^{pre}, \mathbf{h}_V^{pre}] = [h_{|\mathbf{y}_1|}^{pre}, \cdots, h_{|\mathbf{y}_W|}^{pre}] \quad (6)$$

where $h_{|\mathbf{y}_i|}^{pre}$ is the last hidden states of prediction network for $i$th hypothesis $\mathbf{y}_i \in A_t^l$, $\mathbf{h}_S^{pre} = [h_{|\mathbf{y}_1|}^{pre}, \cdots, h_{|\mathbf{y}_{S'}|}^{pre}]$, $\mathbf{h}_V^{pre} = [h_{|\mathbf{y}_{S'+1}|}^{pre}, \cdots, h_{|\mathbf{y}_W|}^{pre}]$, and $S'$ is the length of $\mathbf{h}_S^{pre}$, which changes every $t$-step. The size of $\bar{\mathbf{h}}^{pre}$ is $(W, D)$, where $D$ is the number of hidden units. Note that $\mathbf{h}_S^{pre}$ and $\mathbf{h}_V^{pre}$ are already used for calculating blank probability with respect to $S_{t-1}^p$ (line 9) and $\bar{V}_{t-1}^p$ (line 12), respectively. This reuse is natural as accumulating a blank probability such as $\Pr(\mathbf{y}_i)\Pr(\phi | \mathbf{y}_i, t-1)$ is not relevant to expanding the hypothesis $\mathbf{y}_i$, and thus, its last output label and corresponding last hidden states are maintained after lines 9 and 12. This implies that the last hidden states used for $S_{t-1}^p$ (line 9) and $\bar{V}_{t-1}^p$ (line 12) can be resued at the $t$-step for calculating $\Pr(k | \mathbf{y}_i, t)$, $k \in K \cup \{\phi\}$ with regard to $S_t^p$ (line 9) and $V_t^p$ (line 10). Note that the actual calculation of hidden states is only required in line 12, as discussed in (12).

To match the dimension of the $t$-step hidden states of the encoder network with $\bar{\mathbf{h}}^{pre}$, we duplicate them up to $W$ by introducing the new axis as follows,

$$\mathbf{h}^{enc} = Duplicate(h_t^{enc}, W) = [h_t^{enc}, \cdots, h_t^{enc}], \quad (7)$$

where $h_t^{enc}$ is from (1) and the size of $\mathbf{h}^{enc}$ is $(W, D)$. With $\bar{\mathbf{h}}^{pre}$ and $\mathbf{h}^{enc}$, all posterior probabilities are obtained as

$$\begin{aligned}\mathbf{p} &= Resize(softmax(linear(f^{joint}(\mathbf{h}^{enc}, \bar{\mathbf{h}}^{pre})))) \\ &= [\Pr(\phi | \mathbf{y}_1, t), \cdots, \Pr(k | \mathbf{y}_i, t), \cdots, \Pr(|K| | \mathbf{y}_W, t)]\end{aligned} \quad (8)$$

where *Resize* changes the size of the softmax result from $(W, |K|+1)$ to $(W|K|+W, 1)$.

To obtain the accumulated posterior probabilites $\mathbf{p}'$, the existing ones for $A_t^l$, i.e., $A_t^l = [\Pr(\mathbf{y}_1), \cdots, \Pr(\mathbf{y}_W)]$ are duplicated up to $|K|+1$ to match the dimension of $A_t^p$ with $\mathbf{p}$ so that $\mathbf{p}'$ is obtained by elementwise multiplication between $A_t^p$ and $\mathbf{p}$ as follows:

$$\begin{aligned}\mathbf{A}_t^p &= Resize(Duplicate(A_t^p, |K|+1)) \\ &= [\Pr(\mathbf{y}_1), \Pr(\mathbf{y}_1), \cdots, \Pr(\mathbf{y}_i), \cdots, \Pr(\mathbf{y}_W), \Pr(\mathbf{y}_W)]\end{aligned} \quad (9)$$

$$\mathbf{p}' = \mathbf{A}_t^p \circ \mathbf{p} \quad (10)$$

where the sizes of $\mathbf{A}_t^p$ and $\mathbf{p}'$ are $(W|K|+W, 1)$.

Note that the accumulated posterior probabilities for each decoding step of RNN-T must be ended with a blank probability for *complete probability*. Thus, we divide $\mathbf{p}'$ into $S_t^p$ (line 9) and $V_t^p$ (line 10) as $V_t^p$ must further be processed, as elements in $V_t^p$ are ended with $\Pr(k | \mathbf{y}_i, t)$ as in line 10. The elements of $S_t^p$ are obtained by selecting every $1+j(|k|+1)$ th element from $\mathbf{p}'$, where $j \in \mathbb{N}$. The others are assigned to $V_t^p$.

Before calculating the blank probabilities for $V_t^p$, we prune $V_t^p$ to have top-$W$ *incomplete probabilities* and obtain $V_t^l$ as in line 11, referred to as the local pruning. We found that multiplying blank probabilities with $V_t^p$ slightly affects the

---

**Algorithm 2**: One-Step Constrained Beam Search

1. **Initialize**: $B_0^l = \{\phi\}$; $\Pr(\phi) = 1$; $\bar{\mathbf{h}}^{pre} = f^{pre}(\mathbf{0}, \mathbf{0}')$
2. **for** $t = 1$ **to** $T$ **do**
3. $\quad A_t^l = B_{t-1}^l$
4. $\quad B_t^l = \{\}$
5. $\quad$ **for** $\mathbf{y}_i$ in $A_t^l$ **do**
6. $\quad\quad \Pr(\mathbf{y}_i) \mathrel{+}= \sum_{\hat{\mathbf{y}}} \Pr(\hat{\mathbf{y}}) \Pr(\mathbf{y}_i | \hat{\mathbf{y}}, t)$,
$\quad\quad$ where $\hat{\mathbf{y}} \in pref(\mathbf{y}_i) \cap A_t^l$ and $|\mathbf{y}_i| - |\hat{\mathbf{y}}| \leq \alpha$
7. $\quad$ **end for**
8. $\quad$ Calculate all probabilities for possible expansions of hypotheses,
9. $\quad S_t^p = \{\Pr(\mathbf{y}_i) \Pr(\phi | \mathbf{y}_i, t) | \mathbf{y}_i \in A_t^l\}$
10. $\quad V_t^p = \{\Pr(\mathbf{y}_i) \Pr(k | \mathbf{y}_i, t) | k \in K, \mathbf{y}_i \in A_t^l\}$
11. $\quad V_t^l = $ Select top $W$ hypotheses from $V_t^p$
12. $\quad \bar{V}_t^p = \{\Pr(\mathbf{y}_i') \Pr(\phi | \mathbf{y}_i', t) | \mathbf{y}_i' \in V_t^l - A_t^l\}$
13. $\quad B_t^l = $ Select top $W$ hypotheses from $S_t^p \cup \bar{V}_t^p$
14. **end for**
15. **Return**: $\mathbf{y}^{final}$ with highest $\log \Pr(\mathbf{y}_i) / |\mathbf{y}_i|$ in $B_T^l$

---

order of top-$W$ entries of $V_t^p$, i.e., the order dominantly depends on $\Pr(k | \mathbf{y}_i, t)$. With local pruning, we do not have to calculate blank probabilities for all hypotheses regarding $V_t^p$.

After local pruning, duplication check—the condition of the hypothesis in line 12—is conducted prior calculating the blank probabilities for hypotheses in $V_t^l$ as the elements of $A_t^l$ and $V_t^l$ can be duplicated. For example, let $\{[a,b],[a]\} \subset A^l$; subsequently $[a]$ is expanded to $[a,b]$ in the expansion search, and $[a,b]$ can be included in $V_t^l$ so that $V_t^l \cap A_t^l \neq \{\}$. Without the duplication check, we cannot guarantee that all candidate hypotheses before global pruning in line 13 are unique, i.e., the duplicated hypotheses can be included in $B_t^l$.

To calculate the blank probabilities for expanded hypotheses $\mathbf{y}_i' \in V_t^l - A_t^l$, we newly compute $\bar{\mathbf{h}}_E^{pre}$ as

$$\mathbf{h}_E^{pre} = [h_{|pa(\mathbf{y}_1')|}^{pre}, \cdots, h_{|pa(\mathbf{y}_N')|}^{pre}], \quad \mathbf{y}^{last} = [y_1^{last}, \cdots, y_N^{last}], \quad (11)$$

$$\bar{\mathbf{h}}_E^{pre} = f^{pre}(\mathbf{y}^{last}, \mathbf{h}_E^{pre}), \quad (12)$$

where $N$ is $|V_t^l - A_t^l|$, $y_i^{last}$ is the last label of $\mathbf{y}_i'$ and $pa(\cdot)$ maps the expanded hypothesis to its parent before expansion. Thus, $h_{|pa(\mathbf{y}_i')|}^{pre}$ is one of the elements of $\bar{\mathbf{h}}^{pre}$ in (6). Note that $\bar{\mathbf{h}}_E^{pre}$ will be reused as $\mathbf{h}_V^{pre}$ in (6) at $(t+1)$-step. Also, the *incomplete probability* for the expanded hypothesis $\mathbf{y}_i'$ is $\Pr(\mathbf{y}_i') = \Pr(pa(\mathbf{y}_i'))\Pr(y_i^{last} | pa(\mathbf{y}_i'), t)$ and $\Pr(\mathbf{y}_i') \in V_t^p$. With $\bar{\mathbf{h}}_E^{pre}$ and $\mathbf{h}^{enc}$, the blank probabilities are computed from (8), and we can obtain $\bar{V}_t^p$ for the *complete probabilities* in line 12.

The *complete probabilities* of the final candidate hypotheses $S_t^p \cup \bar{V}_t^p$ are pruned as in line 13 and the final $W$ hypotheses at the $t$-step decoding procedure are saved to $B_t^l$. Note that, $h_{|\mathbf{y}_i|}^{pre}$ such that $\mathbf{y}_i \in A_t^l \cap B_t^l$ will be reused as $\mathbf{h}_S^{pre}$ in (6). The above procedures are conducted for $T$-step, and the final predicted output label sequence is obtained as in line 15.

## IV. EXPERIMENTS AND RESULTS

### A. Experimental Setup

We used the English and Korean speech corpora, TIMIT [13], LibriSpeech [14], and KakaoMini (Korean corpus). The KakaoMini corpus consists of 1.5 M Korean utterances (~ 1000 h), recorded for distant speech recognition for which the recording distance was 1–4 m. The recording was conducted using a KakaoMini smart speaker in reverberant and noisy environments with 5–30 dB SNRs. As an evaluation set, we used the core test set with 192 utterances on the TIMIT corpus and randomly selected 1000 utterances from the LibriSpeech (~2 h) and KakaoMini (~1 h) corpora. As input features, we used 40- and 80-dimensional globally normalized log Mel-filter bank coefficients for TIMIT and the other corpora, respectively.

For the TIMIT corpus, we used 3- and 1-layer LSTMs with 256 cells as encoder and decoder networks, respectively. The joint network had 256 hidden units. All 61 phoneme labels were used during training and decoding. They were mapped to 39 labels for evaluation. We followed the learning strategy in [15].

For the LibriSpeech and KakaoMini corpora, we used 5- and 2-layer LSTMs with 512 cells as encoder and decoder networks, respectively and the joint network with 512 hidden units; we used 256 and 1000 word-pieces for labels [16], respectively. We used the Adam optimizer [17] with an initial learning rate of 0.001; the learning rate scheduling was performed according to the validation set [18], the dropout with a rate of 0.2, SpecAugmentation [19], and layer normalization [20].

The implementation was based on Python3 and used Tensorflow [21]. Although the training was conducted using GPUs (4 Nvidia Tesla-V100), the inference with beam search was performed using a CPU (Intel Xeon Processor Gold 5120).

For baseline methods, the beam search in [7] (B1) and the improved one in [12] (B2) were used. B2 has hyperparameters referred to as "*expand_beam*" and "*state_beam*", which were set to the best ones described in [12]. Note that we did not fully optimize our models by searching numerous hyperparameters as our focus was the decoding speed. Further, we did not perform precision quantization, frame stacking, and apply a time-reduction layer to the encoder network to speed up the beam search as our objective is comparing the OSC beam search speed with baselines from an algorithmic perspective.

As evaluation metrics, PER for the TIMIT corpus, WER for LibriSpeech and KakaoMini corpora, and the real time factor (processing time divided by audio duration) at 90 percentile (RT-90) were used.

*B. Experimental Results and Discussion*

Table II compares the OSC beam search with B1 and B2 for different parameter settings regarding $W$ and $\alpha$. For the TIMIT corpus, the OSC beam search was significantly speedup compared to B1 and B2 in RT-90, even outperforming them in PER. We found that PER improvement is from the duplication check; thus, when disregarding the duplication check, the PER of the OSC beam search with $\alpha$ set to 1 was degraded to 24.42, 24.07, and 24.36 when $W$ is set to 5, 10, and 20, respectively. Note that the absence duplication check cannot guarantee the beam search effect as the duplicated hypotheses can be included in a hypothesis set. Further, increasing $W$ is not proportional to improving PER, which implies that the predicted output sequence with the highest accumulated posterior probabilities does not correspond to the best sequence as the output probability lattice itself is predicted.

As expected, the RT-90 gap obtained by comparing O-$W$-1 with B1-$W$ and B2-$W$ is found to be extended with an increase in $W$, i.e., 2.87 × (B1) and 1.54 × (B2) speed up when $W$ is set to 5, and to 7.24 × (B1) and 3.66 × (B2) speedup when $W$ is set to 20. This result is natural as the OSC beam search removes the while-loop in Algorithm 1. The PER was better when the prefix constraint $\alpha$ is set to 2 than set to 1, as $\alpha$ with 2 covers most prefixes listed in Table I, this implying that a better approximation of the prefix search can be conducted with a higher $\alpha$. Note that, we studied when $\alpha > 2$; however, the effect on both PER and WER was found negligible.

For LibriSpeech and KakaoMini corpora, the RT-90 gap when comparing O-$W$-1 with B1-$W$ and B2-$W$ was further extended compared to using the TIMIT corpus as a model size for that corpora was bigger than that for the TIMIT corpus. Regardless of the corpus type and model size, our main contribution is that the RT-90 of B1 and B2 increased by more than double as the $W$ doubled, whereas the OSC beam search showed less than a doubled RT-90 in most of the cases; thus, the OSC beam search has an algorithmic improvement compared to B1 and B2. Further, WER was degraded when $\alpha$ was set to 1 compared to B1 and B2, although RT-90 was still significantly improved. However, the OSC beam search with $\alpha$ set to 2 outperformed both B1 and B2 in WER and RT-90 across all $W$ settings, implying that a better prefix search approximation should be considered for some cases.

## V. CONCLUSION

We proposed the OSC beam search to accelerate RNN-T inference by vectorizing the multiple hypotheses. The crucial point allowing the vectorization was OSC. We obtained further acceleration from constrained prefix search by pruning the search space that can be redundant. In addition, OSC beam search checks the duplication among hypotheses during decoding process, leading to WER and PER improvement. However, we only focused on vectorizing multiple hypotheses; thus, vectorization of speech utterances will be our future works.

TABLE II
THE PERFORMANCE COMPARISON ON TIMIT, LIBRISPEECH AND KAKAOMINI. THE NUMBERS IN BOLD INDICATES THE BEST RESULT.

| Dataset | Metric | B1-5 | B1-10 | B1-20 | B2-5 | B2-10 | B2-20 | O-5-1 | O-10-1 | O-20-1 | O-5-2 | O-10-2 | O-20-2 |
|---|---|---|---|---|---|---|---|---|---|---|---|---|---|
| TIMIT | PER | 22.32 | 22.76 | 23.73 | 22.30 | 22.88 | 23.86 | 22.25 | 22.10 | 22.21 | 22.16 | **22.04** | 22.06 |
|  | RT-90 | 0.505 | 1.127 | 2.527 | 0.272 | 0.605 | 1.276 | **0.176** | 0.220 | 0.349 | 0.197 | 0.294 | 0.503 |
| Librispeech | WER | 11.18 | 11.3 | 11.62 | 10.99 | 11.2 | 11.77 | 11.75 | 11.52 | 11.36 | 11.54 | 11.13 | **10.92** |
|  | RT-90 | 2.740 | 6.919 | 13.063 | 2.682 | 5.685 | 12.996 | **0.876** | 1.033 | 1.322 | 0.901 | 1.120 | 1.690 |
| KakaoMini | WER | 5.381 | 5.482 | 4.898 | 5.228 | 4.822 | 4.822 | 5.660 | 5.964 | 5.761 | 5.000 | 4.746 | **4.543** |
|  | RT-90 | 7.297 | 16.613 | 43.831 | 4.849 | 10.271 | 24.187 | **0.671** | 1.018 | 1.771 | 1.032 | 1.348 | 2.225 |

The baseline methods (B1 and B2) were described as two dimensions: beam search method (B1 or B2) and $W$ (5, 10 or 20). OSC beam search abbreviated to O was described as three dimensions: beam search method (O), $W$ (5, 10 or 20) and $\alpha$ (1 or 2).